\documentclass[letterpaper, 10 pt, conference]{ieeeconf}  % Comment this line out if you need a4paper

\IEEEoverridecommandlockouts                              % This command is only needed if 
\usepackage[T1]{fontenc}
\usepackage{preamble}
\usepackage{cleveref}
\usepackage{graphicx}
\usepackage{multirow}
\usepackage{booktabs}
\usepackage{pifont}
\newcommand{\fref}[1]{Fig. \ref{#1}}
\newcommand{\tref}[1]{Table \ref{#1}}
\newcommand{\eref}[1]{Eq. \ref{#1}}

\title{\LARGE \bf
LST-SLAM: A Stereo Thermal SLAM System for Kilometer-Scale Dynamic Environments
}

\author{Zeyu Jiang$^{1}$, Kuan Xu$^{2}$, Changhao Chen$^{1*}$
\thanks{$^{1}$Zeyu Jiang and Changhao Chen are with PEAK-Lab, The Hong Kong University of
Science and Technology (Guangzhou), Guangzhou, 511453, China
        {\tt\small zjiang122@connect.hkust-gz.edu.cn, changhaochen@hkust-gz.edu.cn}}%
\thanks{$^{2}$Kuan Xu is with the School of Electrical and Electronic Engineering, Nanyang Technological University, 50 Nanyang Avenue, Singapore 639798
        {\tt\small kuan.xu@ntu.edu.sg}}%
\thanks{$^*$ Corresponding Author. }
}

\begin{document}

\maketitle
\thispagestyle{empty}
% \pagestyle{empty}
% 启用页码显示
\pagestyle{plain}

%%%%%%%%%%%%%%%%%%%%%%%%%%%%%%%%%%%%%%%%%%%%%%%%%%%%%%%%%%%%%%%%%%%%%%%%%%%%%%%%
\begin{abstract}
Thermal cameras offer strong potential for robot perception under challenging illumination and weather conditions. However, thermal Simultaneous Localization and Mapping (SLAM) remains difficult due to unreliable feature extraction, unstable motion tracking, and inconsistent global pose and map construction, particularly in dynamic large-scale outdoor environments.
To address these challenges, we propose LST-SLAM, a novel large-scale stereo thermal SLAM system that achieves robust performance in complex, dynamic scenes. Our approach combines self-supervised thermal feature learning, stereo dual-level motion tracking, and geometric pose optimization.  We also introduce a semantic–geometric hybrid constraint that suppresses potentially dynamic features lacking strong inter-frame geometric consistency. Furthermore, we develop an online incremental bag-of-words model for loop closure detection, coupled with global pose optimization to mitigate accumulated drift. Extensive experiments on kilometer-scale dynamic thermal datasets show that LST-SLAM significantly outperforms recent representative SLAM systems, including AirSLAM and DROID-SLAM, in both robustness and accuracy.
\end{abstract}

%%%%%%%%%%%%%%%%%%%%%%%%%%%%%%%%%%%%%%%%%%%%%%%%%%%%%%%%%%%%%%%%%%%%%%%%%%%%%%%%

\section{Introduction}

Visual Simultaneous Localization and Mapping (SLAM) enables autonomous robots to estimate their motion in real time while simultaneously building a map of their surroundings using cameras. Owing to its balance between cost and functionality, visual SLAM has been widely adopted in various robotic applications, such as service and delivery robots \cite{macario2022comprehensive}. However, RGB camera–based visual SLAM often suffers from severe performance degradation in challenging illumination conditions, such as low-light environments or sudden lighting changes.

Thermal cameras provide a promising alternative. Unlike RGB cameras, which rely on reflected visible light, thermal cameras capture infrared radiation emitted by objects, allowing robust imaging in poor lighting conditions and even in complete darkness. They also demonstrate greater resilience in environments affected by fog, dust, or smoke \cite{vidas2012hand}.
Consequently, thermal camera-based SLAM systems \cite{keil2024towards, wu2023improving} have demonstrated strong potential under illumination-challenging conditions. Nonetheless, most existing thermal SLAM systems rely on feature extraction and tracking methods originally designed for RGB images, despite the fact that thermal imaging is physically different and inherently suffers from low contrast, weak texture, and high noise. Moreover, non-uniformity correction (NUC) introduces inter-frame inconsistencies \cite{borges2016practical}, making it difficult to extract sufficiently stable features for long-term motion tracking. 

\begin{figure}[t]
  \centering
  \captionsetup{font=small}
  \includegraphics[width=\linewidth]{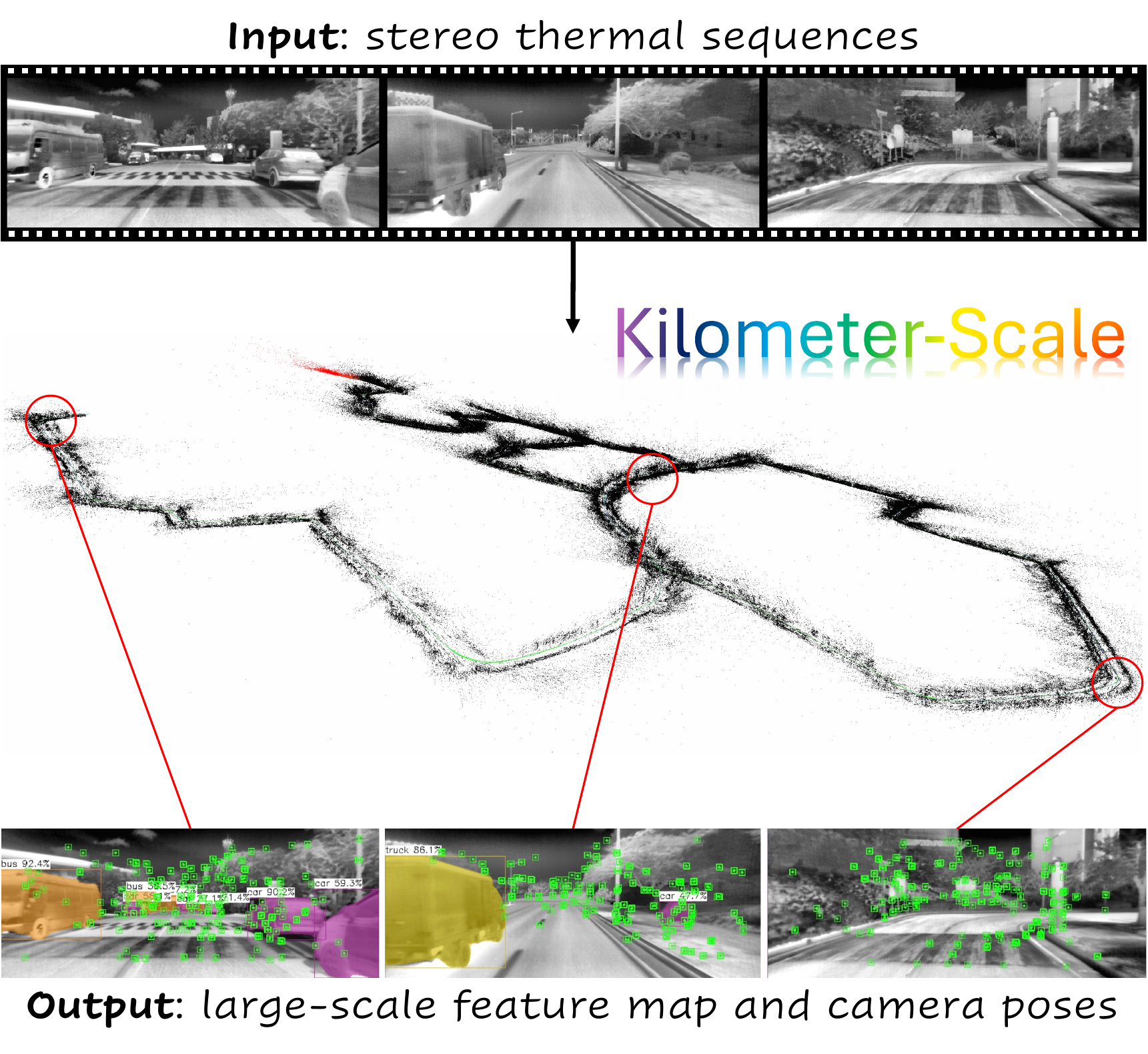}
  \caption{LST-SLAM enables robust localization and mapping in kilometer-scale dynamic thermal scenes. The system leverages self-supervised thermal features and a dual-level tracking strategy, while an incremental online BoW supports loop closure detection and global pose optimization.}
  \label{fig:teaser}
  \vspace{-15pt}
\end{figure}

To address these limitations, several works have fused thermal cameras with additional sensors such as LiDAR \cite{chen2022eil}, RGB cameras \cite{qin2023bvt}, or IMUs \cite{zhao2020tp}. While effective, these approaches increase hardware costs and require substantial calibration and synchronization effort, reducing their practicality and leaving the fundamental challenge of stable thermal-only motion tracking unresolved.
Furthermore, realizing robust thermal SLAM in large-scale outdoor environments remains an underexplored area. Outdoor scenes are typically unstructured, offering fewer stable features than indoor environments, making visual tracking prone to frequent failures. The prevalence of dynamic objects further exacerbates visual aliasing, reducing the robustness of feature tracking. During back-end pose optimization, the outliers introduced by dynamic objects further degrade system accuracy \cite{wu2023improving}. Together, these challenges significantly compromise the robustness and accuracy of existing thermal SLAM systems in large-scale scenarios.

In this work, we propose \textbf{LST-SLAM}, a novel \textbf{L}arge-scale \textbf{S}tereo \textbf{T}hermal \textbf{SLAM} system designed to achieve reliable performance in kilometer-scale, illumination-challenging, and dynamic outdoor environments. LST-SLAM integrates thermal feature self-learning, dual-level motion tracking, and geometric pose optimization. To enable stable thermal feature extraction, we introduce a thermal point network with a thermal homography training loss that adapts geometric knowledge learned in the RGB domain to the thermal domain. To improve motion tracking robustness, we design a stereo dual-level feature-tracking strategy that jointly optimizes low-level photometric and high-level descriptor errors. In addition, we leverage semantic and geometric cues to perform dynamic feature filtering, mitigating the impact of moving objects commonly present in real-world traffic scenes. Finally, we construct an incremental bag-of-words (BoW) of thermal features for loop closure detection, coupled with global pose optimization to reduce accumulated drift. In real-world kilometer-scale experiments, LST-SLAM achieves 75.8\% and 66.8\% lower localization errors than representative systems, i.e., AirSLAM (2025) and DROID-SLAM (2021), respectively.

In summary, our main contributions are as follows:
\begin{itemize}
\item We propose LST-SLAM, a novel thermal SLAM system tailored for large-scale, dynamic, and illumination-challenging outdoor scenarios, featuring thermal feature self-learning, dual-level motion tracking, incremental thermal BoW construction, and geometric pose optimization.
\item We introduce a thermal point network with a thermal homography training loss that adapts geometric knowledge from the RGB domain to the thermal domain. Together with a stereo dual-level tracking strategy and dynamic feature filtering, our system transforms learned thermal features into reliable motion tracking.
\item We conduct extensive experiments in highly challenging kilometer-scale outdoor environments, demonstrating that LST-SLAM achieves superior feature tracking stability and localization accuracy compared to several representative SLAM systems.
\end{itemize}

\section{Related Works}

\noindent\textbf{Thermal Feature Extractor} 
Mouats \etal \cite{mouats2018performance} established a benchmark thermal dataset to evaluate the performance of handcrafted feature extractors. However, traditional detectors and descriptors \cite{lowe2004distinctive, rublee2011orb, yi2016lift}, initially designed for visible-spectrum images, are difficult to adapt to thermal data since they rely heavily on image gradient information. To address these limitations, deep learning approaches have been explored. Zhao \etal \cite{zhao2020tp} proposed ThermalPoint, a supervised method trained on authentic thermal images, which mitigates weak texture and illumination variations while improving noise robustness. Lu \etal \cite{lu2021superthermal} introduced a complete learning pipeline for robust thermal feature detection and description. Deshpande \etal \cite{deshpande2021matching} designed a triplet-based Siamese CNN to extract features from arbitrary thermal images. More recently, Tuzcuoglu \etal \cite{tuzcuouglu2024xoftr} developed a cross-modal, cross-view approach for local feature matching between thermal infrared and visible images. Such cross-modal strategies offer valuable insights for advancing thermal feature learning.

\noindent\textbf{Thermal SLAM} 
Compared to RGB SLAM, thermal SLAM has received relatively limited attention. Most prior efforts have relied on multi-sensor fusion strategies, combining thermal cameras with LiDAR \cite{chen2022eil, shin2019sparse}, IMU \cite{zhao2020tp, saputra2021graph, jiang2022thermal}, or RGB cameras \cite{chen2017rgb, qin2023bvt} to leverage complementary sensing modalities for improved robustness in extreme conditions.
Early work by Vidas \etal \cite{vidas2012hand} presented the first SLAM system using a single thermal sensor, reporting large reprojection errors that highlighted challenges in thermal feature matching. More recently, Wu \etal \cite{wu2023improving} proposed a monocular thermal SLAM method that incorporates dynamic object segmentation to mitigate feature degradation in dynamic environments. Xu \etal \cite{xu2025slam} introduced an end-to-end monocular thermal SLAM system that achieves accurate localization and dense 3D mapping in nighttime conditions. Gupta \etal \cite{keil2024towards} investigated feature matching and loop closure under diurnal temperature variations, proposing a long-term thermal SLAM framework. Wu \etal \cite{wu2025monocular} further integrated Neural Radiance Fields (NeRF) into thermal SLAM for thermal-based 3D reconstruction. These methods have improved RGB-based SLAM systems in specific modules, but they have not been designed for thermal features across the entire system. Therefore, previous research struggles to be applied to kilometer-scale real-world thermal scenarios.

\begin{figure*}[t]
  \centering
  \vspace{5pt}
  \captionsetup{font=small}
  \includegraphics[width=\linewidth]{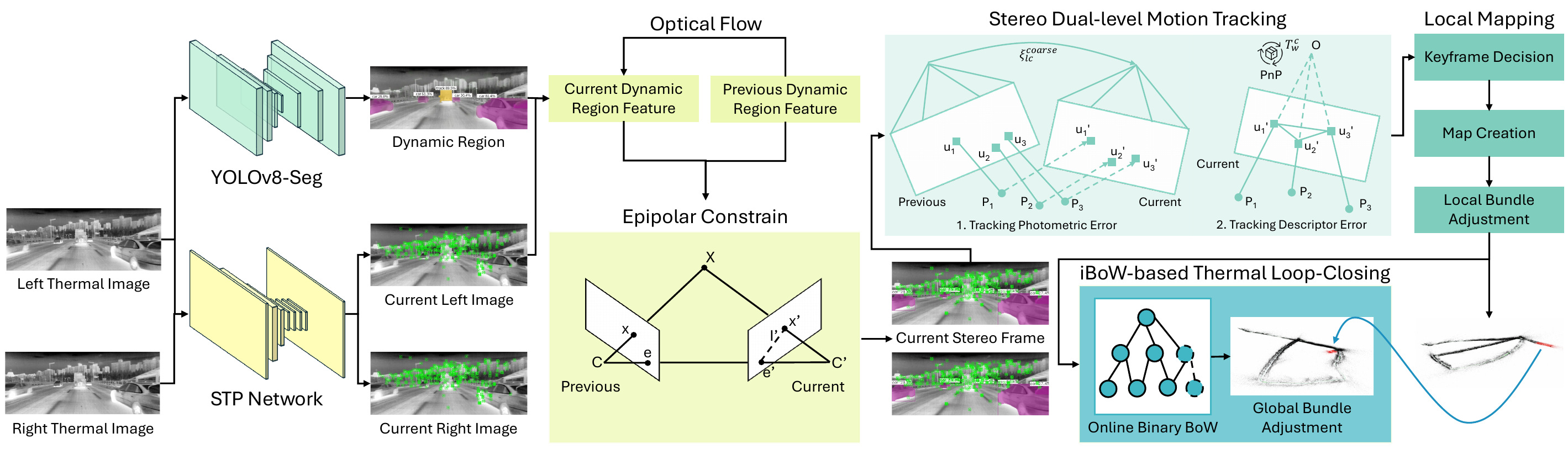}
  \caption{LST-SLAM system pipeline. Stereo inputs are processed with self-supervised feature point and dynamic filtering networks to obtain robust thermal features, which drive tracking, mapping, and local optimization. Binarized descriptors build an incremental BoW (iBoW) for loop closure and global pose optimization.}
  \label{fig:DST-SLAM}
  \vspace{-15pt}
\end{figure*} 

\section{Methodology}

As illustrated in \fref{fig:DST-SLAM}, LST-SLAM combines a learning-based thermal feature detection module with dynamic object filtering, an incremental online BoW framework for loop closure, and a pose optimization backend. Pre-processed thermal images (Section~\ref{Method: Thermal Preprocess}) are first fed into the self-supervised thermal point network described in Section~\ref{Method: Feature Detection}. Robust stereo static thermal feature keypoints are then obtained via the dynamic object filtering model (Section~\ref{Method: Dynamic Object Filtering}). These keypoints are used in stereo dual-level motion tracking (Section~\ref{Method: Stereo Dual-level Tracking}) and bundle adjustment (BA) (Section~\ref{Method: Local Map Optimization}) to refine both camera poses and the reconstructed map. Finally, during loop closure (Section~\ref{Method: Deep Thermal Point Loop Closing}), the incremental online BoW scheme enables generalized loop detection and global pose optimization.

\subsection{Thermal Image Preprocess}
\label{Method: Thermal Preprocess}

Raw infrared images are typically 16-bit, low-contrast, and noisy, making them unsuitable for direct feature extraction. To address this, we apply a preprocessing pipeline to enhance image quality before subsequent processing. Specifically, each raw 16-bit thermal frame is first normalized to 8-bit using percentile stretching, with intensity bounds defined by the $1\%$ and $99\%$. This suppresses outliers while preserving the main scene content.
To reduce flickering caused by frame-to-frame variations, the percentile bounds are smoothed over time using an exponential moving average (EMA):
\begin{equation}
l_t = \alpha l_{t-1} + (1-\alpha)\hat{l}_t,\quad 
h_t = \alpha h_{t-1} + (1-\alpha)\hat{h}_t
\end{equation}
where $\hat{l}_t$ and $\hat{h}_t$ denote the raw percentile estimates, and $\alpha \in [0,1)$ is the smoothing factor. We observed negligible performance differences in the range $\alpha \in [0.7, 0.9]$. The stretched image is then linearly mapped to the range $[0,255]$.

Finally, contrast-limited adaptive histogram equalization (CLAHE) is applied to enhance local contrast, with the clip limit and tile size chosen to balance detail enhancement against noise amplification. This preprocessing pipeline produces temporally stable, high-contrast thermal images that are well-suited for both visualization and downstream computer vision tasks.

\subsection{Self-Supervised Thermal Feature Learning}
\label{Method: Feature Detection}

In feature-based visual SLAM, the quality of feature descriptors is critical for data association. High-quality descriptors enable reliable feature matching, which directly impacts tracking stability and loop closure success. Conversely, poor descriptors often lead to mismatches, resulting in pose estimation errors and degraded map consistency. Recent advances in self-supervised feature learning, such as SuperPoint \cite{detone2018superpoint}, have demonstrated the effectiveness of deep neural networks trained on large datasets for extracting robust keypoints in challenging scenarios. However, directly applying learning-based feature extractors, such as SuperPoint, to thermal images leads to cumulative localization errors and frequent tracking failures due to the limited number of reliable keypoints. Our experiments (see \fref{fig:curve}) further reveal that SuperPoint produces feature matches in the thermal domain with limited robustness.

\begin{figure}[t]
  \vspace{5pt}
  \centering
  \captionsetup{font=small}
  \includegraphics[width=\linewidth]{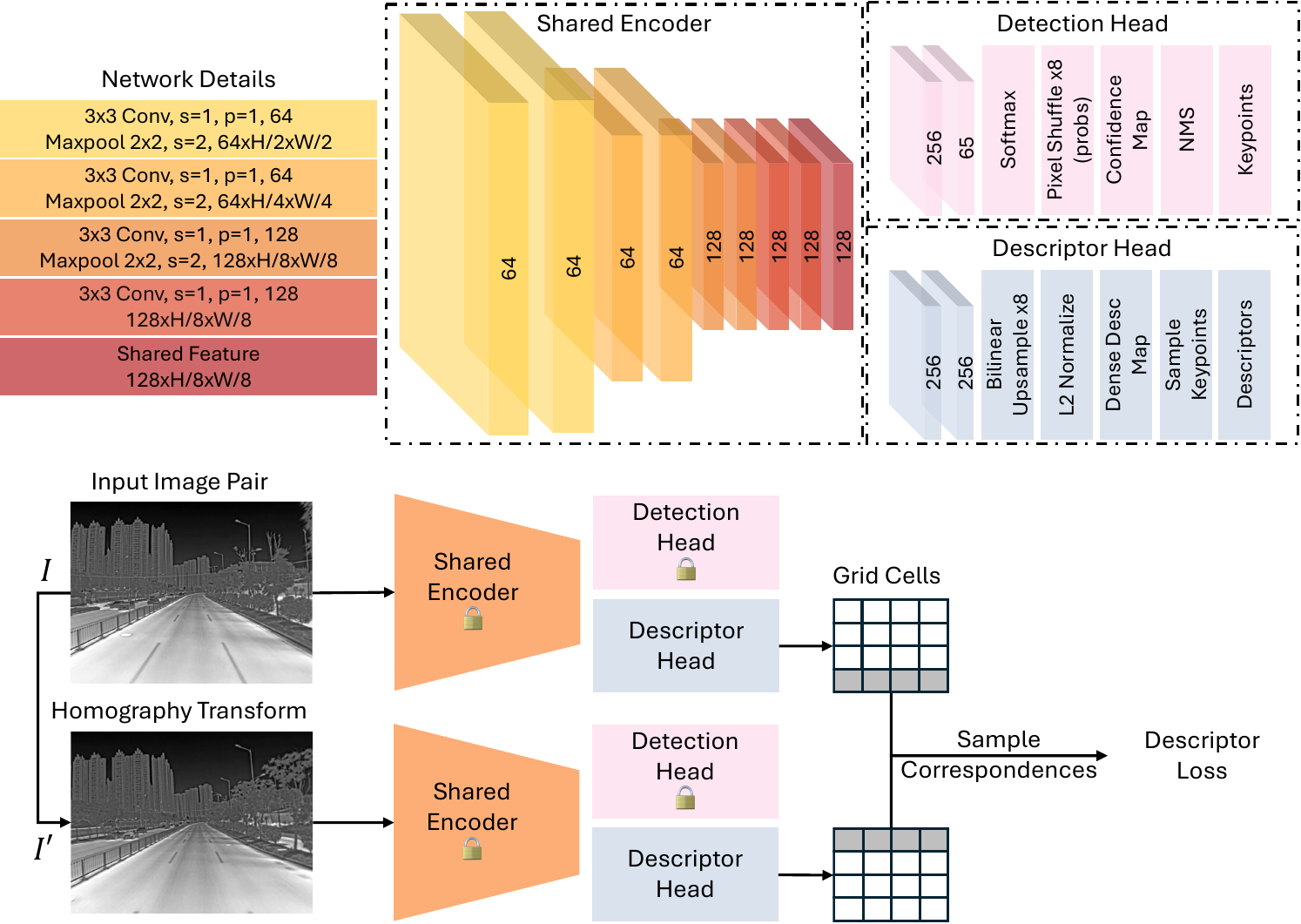}
  \caption{\textbf{Top:} Architecture and details of the STP network. \textbf{Bottom:} Schematic diagram of adaptive self-supervised training in STP networks under thermal modality.}
  \label{fig:STP}
  \vspace{-15pt}
\end{figure}

To address this, we introduce a dedicated solution for thermal feature learning, termed the Self-supervised Thermal Point (STP) network. As illustrated in \fref{fig:STP}, STP leverages descriptor distances between homography-transformed thermal image pairs as self-supervised training signals, enabling effective adaptation to the thermal domain. To address the scarcity of thermal training data, STP is initialized with weights from the SuperPoint, thereby transferring geometric priors learned in the RGB domain to thermal feature learning.

\noindent\textbf{STP Network Architecture}\; Following SuperPoint architecture, our network consists of a shared encoder for high-level feature abstraction and two decoders, as illustrated in \fref{fig:STP}. The detection head predicts the spatial probability distribution of interest points, producing a tensor $\mathcal{C} \in \mathbb{R}^{65 \times \frac{H}{8} \times \frac{W}{8}}$, where the 65 channels represent an $8 \times 8$ grid cell plus an additional dustbin channel. Reshaping this tensor yields an $H \times W$ heat map of pixel-level saliency. The descriptor head generates $\mathcal{D} \in \mathbb{R}^{256 \times \frac{H}{8} \times \frac{W}{8}}$, which is upsampled via bicubic interpolation to $\mathbb{R}^{256 \times {H} \times {W}}$. Each descriptor vector is then L2-normalized to unit length. After filtering the detected feature points, their corresponding descriptors are extracted.

\noindent\textbf{Thermal Homography Training Loss}\; We freeze the shared encoder and detection head to preserve keypoint localization, while training only the descriptor head to enhance feature representation in thermal datasets. This design obviates the need for pseudo-labels. For each input image, a second view is generated by sampling a homography transform with independent translation, rotation, scaling, and perspective components, ensuring a patch overlap ratio above 0.85. Photometric augmentations are applied to both views, including random brightness and contrast adjustments, additive Gaussian noise, speckle noise, and motion blur.

Given a pair of gray-scale images related by a homography $H$, the grid coordinates are $i = (h_i, w_i) \in \{0,\ldots,H_c-1\}\times\{0,\ldots,W_c-1\}, \; j = (h_j, w_j)$, where $H_c=\frac{H}{8}$ and $W_c=\frac{W}{8}$ in all our experimental setting. We defined $\mathbf{d}_i\in\mathbb{R}^{256}$ and $\mathbf{d}_j^{\prime}$ denote the L2-normalized descriptors at the centers of cell $i$ in the reference image and cell $j$ in the warped image, respectively. $\boldsymbol{\Delta}_{ij}$ is the Euclidean distance between the two cell centres after alignment by $H$. We use the following constants $r_+=8\mathrm{~px}, m_p=1, m_n=0.2, \lambda_d=250.$ Then we calculate the positive and negative correspondence matrix by 
$s_{ij}=1$ if $\Delta_{ij}\le r_{+}$, and $0$ otherwise. 
The similarity is represented as 
$\boldsymbol{s}_{ij}\in\{0,1\}$ and 
$c_{ij}=\max(0,\mathbf{d}_i^\top \mathbf{d}_j^{\prime})\in[0,1]$.
The descriptor loss function is defined as
\begin{equation}
\begin{aligned}
\mathcal{L}_{\mathrm{ij}}
&= \lambda_d s_{ij}(m_p-c_{ij})_+ + (1-s_{ij})(c_{ij}-m_n)_+, \\
\mathcal{L}_{\mathrm{d}}
&= \frac{1}{N}\sum_{i=1}^{H_cW_c}\sum_{j=1}^{H_cW_c} \mathcal{L}_{\mathrm{ij}}.
\end{aligned}
\end{equation}
with $(x)_+=\max(0,x)$, $N=B(H_cW_c)^2$ denoting the total number of candidate pairs where $B$ denotes batch size. 

\subsection{Dynamic Feature Point Filtering}
\label{Method: Dynamic Object Filtering}

In real-world scenarios, feature points originating from dynamic objects can severely degrade motion tracking performance. This issue is particularly critical in thermal SLAM, where reliable features are already difficult to extract and match. To this end, we design a semantic-geometric hybrid constraint that suppresses potentially dynamic features when they lack strong inter-frame geometric consistency. This strategy enhances feature set reliability and provides a more robust foundation for subsequent pose estimation.

To balance real-time performance with detection accuracy, we integrate a YOLOv8-Seg network \cite{redmon2016you} into the SLAM front-end. Although YOLOv8-Seg is originally trained on RGB datasets, we observe that dynamic categories exhibit strong structural and contour cues in thermal imagery, enabling robust detection even without retraining. The network supports 80 object categories \cite{lin2014microsoft}, and moving entities such as cars, buses, and trucks are designated as dynamic classes based on scene characteristics. Each input image is processed with non-maximum suppression (NMS) and mask post-processing to produce pixel-level instance masks, which are then merged into frame-level dynamic masks.

The proposed strategy is embedded into the tracking thread, ensuring minimal overhead, where keypoints within dynamic regions undergo additional optical flow and geometric consistency checks. Inter-frame correspondences are first obtained using a sparse optical flow tracker $\mathcal{T}:\mathbf{p}_t\mapsto\mathbf{p}_{t-1}$, where $\mathbf{p}_t=(u,v,1)^\top$ denotes a keypoint in the current frame. To estimate the underlying epipolar geometry, we select correspondences outside the semantic dynamic mask and compute the fundamental matrix $F_{t,t-1}$ using RANSAC: $\mathbf{p}_t^\top F_{t,t-1}\mathbf{p}_{t-1}=0$. Given $\mathrm F_{t,t-1}$, each correspondence $(\mathbf{p}_t,\mathbf{p}_{t-1})$ is evaluated via its point to epipolar line distance. The epipolar line in the previous frame can be written as $\mathbf{l}_{t-1}=F_{t,t-1}\mathbf{p}_t=(a,b,c)^\top,$ and the geometric error takes the form
\begin{equation}
d(\mathbf{p}_{t-1},\mathbf{l}_{t-1})=\frac{|au_{t-1}+bv_{t-1}+c|}{\sqrt{a^2+b^2}}.
\end{equation}
Keypoints inside semantic dynamic regions are retained only if their epipolar distance satisfies $d(\mathbf{p}_{t-1},\mathbf{l}_{t-1})<\tau,$ with a strict threshold $\tau=0.5$ pixels in our implementation. In contrast, keypoints outside dynamic masks are preserved as long as the optical flow tracking is successful.

\subsection{Stereo Dual-level Motion Tracking}
\label{Method: Stereo Dual-level Tracking}
For stereo triangulation, left-right keypoint correspondences $(u_L,v_L)$ and $(u_R,v_R)$ are first obtained by Hamming matching. The disparity is $d_i = u_L - u_R$, yielding the depth $Z_i = \tfrac{bf}{d_i}$, where $b$ and $f$ denote the baseline and focal length, respectively. The corresponding 3D point in the camera frame is then given by $\mathbf{p}_i^c = Z_i K^{-1}[u_L, v_L, 1]^\top$. This yields metrically scaled depth at the frame rate.
If a previous pose $\mathbf{T}^{l}_{w}$ and velocity $\mathbf{V} \approx \mathbf{T}^{l}_{w} \left(\mathbf{T}^{l-1}_{w}\right)^{-1}$ exist, the current pose is predicted as $\hat{\mathbf{T}}^{c}_{w} = \mathbf{V}\mathbf{T}^{l}_{w}$.
Our stereo tracking framework combines low-level photometric alignment with high-level descriptor-based matching:

% \noindent\textbf{Low-level photometric tracking}\
% We estimate the coarse relative pose $\boldsymbol{\xi}_{lc}^{\mathrm{coarse}}$ by minimizing the patch‑based loss, following the semi-direct VO formulation \cite{forster2014svo}:
% \begin{equation}\label{eq: 5}
% \boldsymbol{\xi}_{lc}^\mathrm{coarse}=\arg\min_{\boldsymbol{\xi}}\frac{1}{2}\sum_{i\in\chi}\left[I_c\left(\pi(e^{\boldsymbol{\xi}}\mathbf{p}_i^l)\right)-I_l\left(\pi(\mathbf{p}_i^l)\right)\right]^2,
% \end{equation}
% where $\pi(\cdot)$  projects a 3D map point to pixels and ${\mathbf{\chi}}$ indexes the keypoints.

\noindent\textbf{Low-level photometric tracking}\
We estimate the coarse relative pose $\boldsymbol{\xi}_{lc}^{\mathrm{coarse}}$ by minimizing a patch-based photometric loss, following the semi-direct VO formulation \cite{forster2014svo}:
\begin{equation}\label{eq: 5}
\boldsymbol{\xi}_{lc}^\mathrm{coarse}
=
\arg\min_{\boldsymbol{\xi}}
\frac{1}{2}
\sum_{i\in\chi}
\left[
I_c\!\left(\pi\!\left(e^{\boldsymbol{\xi}}\mathbf{p}_i^l\right)\right)
-
I_l\!\left(\pi\!\left(\mathbf{p}_i^l\right)\right)
\right]^2 ,
\end{equation}
where $I_l(\cdot)$ and $I_c(\cdot)$ denote the image intensity functions of the last frame $l$ and the current frame $c$, respectively; 
$\mathbf{p}_i^l$ is the 3D point expressed in the coordinate frame of $l$; 
$\pi(\cdot)$ denotes the camera projection model; 
and $\chi$ indexes the selected keypoints.

% \noindent\textbf{High-level descriptor tracking}\
% With the coarse pose, every active map point is re-projected and matched within a radius $r$ satisfying $\lVert\mathbf{u}_j-\pi(\hat{\mathbf{T}}_w^c\,\mathbf{p}_i^w)\rVert \le r$ and $\mathrm{dist}_H(\mathbf{d}_i,\mathbf{d}_j) \le \tau_H$. Here, $\mathrm{dist}_H(\mathbf{d}_i,\mathbf{d}_j)$ denotes the \textit{Hamming distance} between the binary descriptors of points $i$ and $j$, and $\tau_H$ is the corresponding matching threshold. The constant coefficient $\rho$ is set to $\frac{1}{2}$. Inlier correspondences $\mathcal I$ drive a PnP-LM:
% \begin{equation}
% \mathbf{T}_w^c=\arg\min_\mathbf{T}\sum_{i\in\mathcal{I}}\rho\left(\left\|\mathbf{u}_i-\pi(\mathbf{T}\mathbf{p}_i^w)\right\|^2\right).
% \end{equation}
% The optimization is implemented using g2o \cite{kummerle2011g} and typically converges within 1 ms.

\noindent\textbf{High-level descriptor tracking}\
With the coarse pose, each active map point is re-projected and matched within a radius $r$ satisfying
$\lVert\mathbf{u}_j-\pi(\hat{\mathbf{T}}_w^c\,\mathbf{p}_i^w)\rVert \le r$.
Although SuperPoint originally outputs \emph{real-valued} descriptors trained with an $\ell_2$ loss, in our system we apply a lightweight \emph{binarization} to enable efficient Hamming-based matching.
Concretely, for each keypoint we sample the SuperPoint descriptor $\mathbf{d}\in\mathbb{R}^{256}$, $\ell_2$-normalize it, and convert it to a 256-bit binary string by a sign test:
$b_k = \mathbb{I}(d_k>0)$.
The resulting bits are packed into 32 bytes, and we compute the Hamming distance
$\mathrm{dist}_H(\mathbf{b}_i,\mathbf{b}_j)\le \tau_H$
as the descriptor similarity test. Inlier correspondences $\mathcal I$ then drive a PnP-LM refinement:
\begin{equation}
\mathbf{T}_w^c=\arg\min_\mathbf{T}\sum_{i\in\mathcal{I}}\rho\left(\left\|\mathbf{u}_i-\pi(\mathbf{T}\mathbf{p}_i^w)\right\|^2\right),
\end{equation}
where $\rho$ is the robust penalty (we use $\rho(x)=\tfrac{1}{2}x$ in our implementation).
The optimization is implemented using g2o \cite{kummerle2011g} and typically converges within 1 ms.

\subsection{Thermal Keyframe Insertion and Bundle Adjustment}
\label{Method: Local Map Optimization}

A new thermal keyframe (KF) is created under either of the following conditions: (i) the temporal gap since the last KF exceeds a predefined maximum threshold, or (ii) the current inlier count drops below roughly three-quarters of that in the reference KF, indicating weakened geometric support. For each keypoint with a reliable depth estimate $0 < Z_i < Z_{\mathrm{th}}$, where $Z_{\mathrm{th}} = \frac{bf}{d_{\mathrm{min}}}$, a new 3D point $\mathbf{p}_i^{w}$ is initialized from \eref{eq: 5} and inserted into the map with an observation edge linked to the new KF. 

%For local bundle adjustment in an asynchronous thread, the local window${\text{current KF}}\cup\bigl(\text{top-}k\text{ covisible KFs}\bigr)$ is jointly refined:

Local BA is asynchronously performed over a sliding window consisting of the current KF and its top-$k$ covisible KFs:
\begin{equation}
\{\mathbf{T},\mathbf{p}\}^\star=\arg\min\sum_{(\mathsf{KF},i)}\rho(\|\mathbf{u}_i^{\mathsf{KF}}-\pi(\mathbf{T}_w^{\mathsf{KF}}\mathbf{p}_i^w)\|^2).
\end{equation}
%Variables associated with outer KFs are Schur-marginalized, and the reduced system is subsequently solved using the Levenberg-Marquardt method.
where $\rho(\cdot)$ denotes a robust loss function. Variables associated with outer KFs are Schur-marginalized, and the reduced system is solved using the Levenberg–Marquardt algorithm.

\subsection{Incremental BoW Construction for Thermal Loop-Closing}
\label{Method: Deep Thermal Point Loop Closing}

\noindent\textbf{Incremental Construction of Binary BoW}\; Traditional BoW models, such as DBoW2 \cite{GalvezTRO12}, require extensive offline training and lack cross-modal generalization. This limitation is particularly severe in the thermal domain, where cameras with different specifications often produce images with inconsistent feature distributions. To address this issue, we adopt an iBoW strategy \cite{garcia2018ibow}, in which the visual dictionary grows online as new descriptors are observed.

For each incoming frame $t$ with descriptors $\mathbf{D}_t = { \mathbf{d}_i^t }$, each descriptor $\mathbf{d}_i^t$ is assigned to the nearest visual word
\begin{equation}
    \mathbf{w}^\ast = \arg\min_{\mathbf{w}_m \in \mathcal{V}} \; \text{Ham}(\hat{\mathbf{d}}_i^t, \mathbf{w}_m)
\end{equation}
if the Hamming distance is below a threshold $\tau$. Otherwise, a new word is created from $\hat{\mathbf{d}}_i^t$. In this way, the vocabulary $\mathcal{V}$ is incrementally constructed from the test sequence itself, eliminating the need for offline training.

Since deep descriptors are high-dimensional feature vectors, we use binary hashing to enable efficient Hamming distance search. Each descriptor component $d_i$ is binarized as
\begin{equation}
\hat{d}_i =
\begin{cases}
1 & d_i \geq 0, \\
0 & d_i < 0,
\end{cases}
\quad i = 1,\ldots,256,
\end{equation}
where $\hat{\mathbf{d}}$ is the resulting binarized descriptor. This simple yet effective transformation preserves the discriminative structure of the feature space while making descriptors compatible with iBoW indexing.

\noindent\textbf{Thermal Loop-Closure Detection}\; 
To mitigate long-term drift and maintain global consistency, we employ a two-stage loop closure pipeline consisting of appearance-based retrieval and geometric verification. Given the binarized descriptors $\hat{\mathbf{D}}_t = \{ \hat{\mathbf{d}}_i^t \}$ from the current frame, the iBoW index retrieves a ranked set of candidate keyframes (KFs). The similarity score between frame $t$ and candidate $j$ is defined as
\begin{equation}
s_{tj} = \frac{1}{|\mathcal{M}_{tj}|} \sum_{(i,k)\in \mathcal{M}_{tj}}
\bigl( 1 - \tfrac{1}{L}\text{Ham}(\hat{\mathbf{d}}_i^t, \hat{\mathbf{d}}_k^j) \bigr),
\end{equation}
where $\mathcal{M}_{tj}$ is the set of tentative matches and $L$ is the descriptor length. To improve temporal robustness, candidates are grouped into islands $\mathcal{I}$ of size $L$, and the best hypothesis is selected as
\begin{equation}
\mathcal{I}^\ast = \arg\max_{\mathcal{I}}
\frac{1}{|\mathcal{I}|} \sum_{j\in \mathcal{I}} s_{tj}.
\end{equation}
The best candidate KF $k^\ast$ undergoes geometric validation by estimating the fundamental matrix $\mathbf{F}$ with RANSAC and counting inliers consistent with epipolar geometry. If the inliers number exceeds $\tau_{\text{inl}}$, a 7-DoF similarity transform $\mathbf{S} = \{s \mathbf{R}|\mathbf{t}\} \in \mathrm{Sim}(3)$ is computed such that $\mathbf{x}_i^t \simeq s \mathbf{R} \mathbf{x}_i^{k^\ast} + \mathbf{t}$, where $s \in \mathbb{R}^+$, $\mathbf{R} \in SO(3)$, and $\mathbf{t} \in \mathbb{R}^3$. This transform is refined through non-linear optimization by minimizing the reprojection error. After refinement, correspondences are re-evaluated, and the number of consistent inliers is checked. If the inlier count exceeds a threshold, the loop closure is accepted into the pose graph; otherwise, the hypothesis is discarded.

% \begin{figure}[t]
%   \centering
%   \captionsetup{font=small}
%   \includegraphics[width=\linewidth]{img/curve10.pdf}
%   \caption{The STP network significantly outperforms other methods and substantially enhances the SuperPoint behavior in terms of the number of matches and inlier points between different frames.}
%   \label{fig:curve}
%   \vspace{-15pt}
% \end{figure}

\noindent\textbf{Global Pose Optimization}\; Once validated, the loop adds a new constraint to the pose graph $\mathcal{G}=(\mathcal{K},\mathcal{E})$. 
We optimize over Sim(3) poses $\{\mathbf{S}_k\}$ by solving
\begin{equation}
\min_{\{\mathbf{S}_k\}} 
\sum_{(i,j)\in \mathcal{E}}
\rho\!\left(
\bigl\| \log_{\mathrm{Sim}(3)} \bigl( \mathbf{S}_{ij}^{-1}\mathbf{S}_i\mathbf{S}_j^{-1} \bigr) \bigr\|^2
\right),
\end{equation}
where $\mathbf{S}_{ij}$ is the measured relative Sim(3) constraint and $\rho(\cdot)$ is a robust loss. Finally, the global BA jointly refines all camera poses $\{\mathbf{T}_k\}$ and landmarks $\{\mathbf{X}_m\}$:
\begin{equation}
\min_{\{\mathbf{T}_k\}, \{\mathbf{X}_m\}}
\sum_{k,m} \rho\!\left(
\| \pi(\mathbf{T}_k \mathbf{X}_m) - \mathbf{u}_{km} \|^2
\right),
\end{equation}
in this expression $\pi(\cdot)$ is the perspective projection and $\mathbf{u}_{km}$ the observed keypoint.

\begin{figure*}[t]
  \centering
  \vspace{5pt}
  \captionsetup{font=small}
  \includegraphics[width=0.95\linewidth]{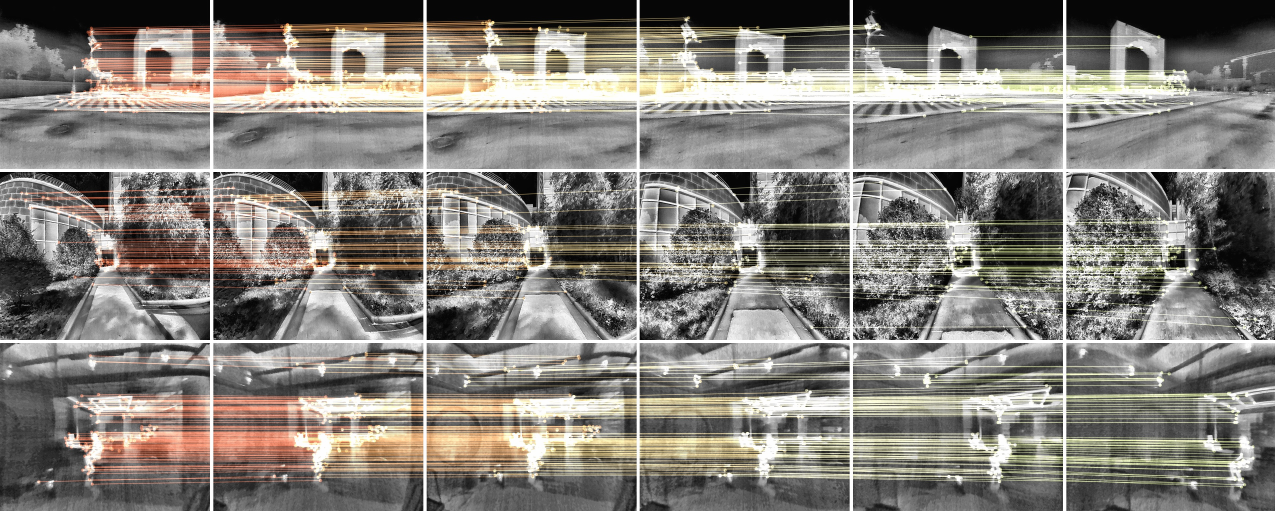}
  \caption{Qualitative results of feature extraction and tracking on the M2DGR dataset. Each row visualizes feature matches over continuous video segments sampled at larger frame intervals. The three sequences were chosen to represent different motion directions.}
  \label{fig:kpt}
  \vspace{-15pt}
\end{figure*}

% In this section, we present the experiment results. In Section \ref{Method: Robustness Evaluation of Feature Extraction} on feature points, we primarily evaluate the tracking robustness of STP network descriptors. In Section \ref{Method: Robustness and Accuracy Comparison of SLAM systems} of the SLAM experiments, we compare numerous SOTA visual SLAM systems. In Section \ref{Method: Ablation Study on DST-SLAM}, we conduct ablation studies on our improved SLAM module.

%In this section, we conducted extensive experiments to evaluate our proposed thermal SLAM system. In Section \ref{Method: Robustness Evaluation of Feature Extraction}, we evaluate the performance of the proposed STP network and feature tracking algorithm. In Section \ref{Method: Robustness and Accuracy Comparison of SLAM systems}, we assess the overall localization accuracy and robustness of the SLAM system by comparing it with other SOTA approaches. In Section \ref{Method: Ablation Study on DST-SLAM}, we conduct ablation studies to validate the contribution of each proposed module to the system. We use two platforms in the experiments. Most evaluations are performed on a personal computer with an AMD Ryzen 9 5900hx CPU and an NVIDIA GeForce RTX 3080 GPU. Due to memory limitations, we were unable to run DROID-SLAM \cite{teed2021droid} and MASt3R-SLAM \cite{murai2025mast3r} on kilometer-scale scenes. Instead, we tested this method using a server equipped with an Intel Xeon Platinum 8358P CPU and four NVIDIA A800 GPUs (80 GiB VRAM).

\section{Experiments}
We conducted extensive experiments to evaluate the LST-SLAM system.
Section \ref{Method: STP Network Training Details} provides training details and experiment setups.
Section \ref{Method: Robustness Evaluation of Feature Extraction} analyzes the performance of the STP network and feature tracking.
Section \ref{Method: Robustness and Accuracy Comparison of SLAM systems} compares the overall localization accuracy and robustness of our SLAM system against representative classic and learning-based methods.
Section \ref{Method: Ablation Study on DST-SLAM} presents ablation studies to assess each module.

\subsection{Experimental Setup}
\label{Method: STP Network Training Details}

\noindent\textbf{STP Network Training Details} The STP network was trained on a large-scale in-vehicle thermal dataset captured with a MAGNITY camera at $640 \times 480$ resolution. The dataset covers 11 urban road segments recorded at different times of day, with additional test frames collected from 5 unseen segments.
Training was performed using the Adam optimizer \cite{kingma2015adam} with an initial learning rate of $1\times10^{-5}$. The network was trained for 300 epochs with a batch size of 2, and the learning rate was decayed to 30\% of its value at epochs 10, 20, 50, 100, and 150. For each epoch, 10,000 images were randomly sampled from a multi-million-frame corpus. Performance was evaluated on 500 validation images using recall and precision. Training on a single NVIDIA GeForce RTX 3080 required approximately three days.

\noindent\textbf{Thermal SLAM Evaluation} Thermal SLAM experiments were conducted on two platforms. Most evaluations ran on a personal computer with an AMD Ryzen 9 5900HX CPU and an NVIDIA GeForce RTX 3080 GPU. Due to memory constraints, kilometer-scale evaluations of DROID-SLAM \cite{teed2021droid} and MASt3R-SLAM \cite{murai2025mast3r} were carried out on a server with an Intel Xeon Platinum 8358P CPU and four NVIDIA A800 GPUs (80 GiB VRAM each).

\subsection{Thermal Feature Extraction Evaluation}
\label{Method: Robustness Evaluation of Feature Extraction}

\noindent\textbf{Datasets \& Baseline}\; For robustness evaluation, we employed the M2DGR dataset \cite{yin2021m2dgr}, which provides $640 \times 512$ thermal streams recorded with a PLUG 617 camera. Four feature extractors were evaluated on identical sequences: STP (ours), SuperPoint \cite{detone2018superpoint}, ORB \cite{rublee2011orb}, and SIFT \cite{lowe2004distinctive}. %We trained the STP network on an extensive in‑vehicle thermal dataset. Frames were captured with a MAGNITY thermal camera at 10 Hz and $640 \times 480$ pixels. The acquisition platform traversed 11 urban road segments at multiple times of day, yielding 85,091 thermal images for training. The test set comprises 20,495 additional images collected over five distinct segments, constituting a large-scale thermal dataset. For the robustness evaluation of feature points, we utilize the M2DGR dataset \cite{yin2021m2dgr}. The thermal stream is recorded using a PLUG 617 camera at a resolution of $640 \times 512$. We selected four widely used feature extractors—STP (Ours), SuperPoint \cite{detone2018superpoint}, ORB \cite{rublee2011orb}, and SIFT \cite{lowe2004distinctive}—run on the identical frame sequences.

\noindent\textbf{Results \& Analysis}\; We assessed feature robustness in terms of tracking quality using two metrics: (i) the number of matches between consecutive frames, and (ii) the number of inlier matches. Putative matches were obtained via mutual nearest-neighbor descriptor search, and inliers were determined using RANSAC \cite{fischler1981random} with essential matrix estimation. For evaluation, we randomly sampled 50 subsequences from 3 scenes. Each subsequence starts from a random frame, followed by 100 consecutive frames, producing 100 image pairs with progressively larger viewpoint changes.

%In evaluating feature-point robustness, we emphasize their effectiveness for tracking. Accordingly, we adopt two metrics: (i) the number of matches between consecutive frames and (ii) the number of available inlier points. The putative matches obtained by mutual nearest-neighbor descriptor search. The inlier matches are received after a RANSAC \cite{fischler1981random} estimate of the essential matrix $E$. We randomly sample 50 subsequences from 3 scenes; for each, a random starting frame is chosen and the subsequent 100 frames in timestamp order form 100 consecutive image pairs, whose viewpoint offset increases with camera motion.

\begin{figure}[t]
  \centering
  \captionsetup{font=small}
  \includegraphics[width=\linewidth]{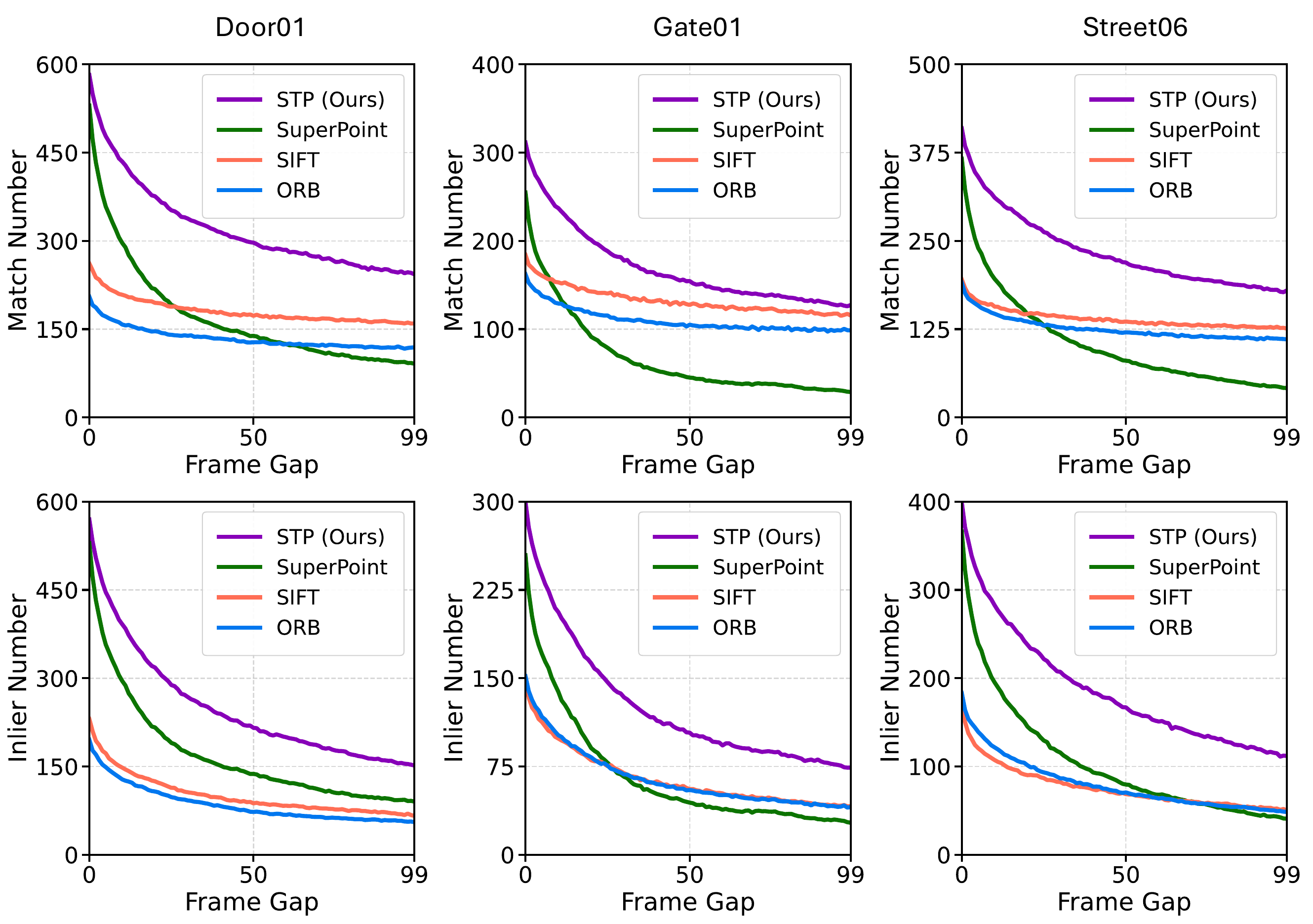}
  \caption{The STP network significantly outperforms other methods and substantially enhances the SuperPoint behavior in terms of the number of matches and inlier points between different frames.}
  \label{fig:curve}
  \vspace{-15pt}
\end{figure}

As shown in \fref{fig:curve}, STP consistently achieves more matches and inliers than all baselines. This result indicates that the extracted thermal feature points are both denser and more robust. Notably, SuperPoint often underperforms traditional methods when frame gaps increase, whereas STP markedly improves SuperPoint’s robustness in the thermal domain. In terms of overall performance curves, our method yields improvements of 35.8\%, 47.5\%, and 48.4\% over SuperPoint, SIFT, and ORB, respectively, in the number of inlier points. These results confirm that self-supervised training on thermal data significantly strengthens feature extraction for thermal SLAM. Visualized matches from STP are presented in \fref{fig:kpt}.

\begin{table*}[htbp]
\centering
\captionsetup{font=small}
\caption{Performance comparison of different odometry and SLAM systems on the MS$^{2}$ dataset across various sequences. The average is calculated over all sequences. The best results are in \textbf{bold}.}
\vspace{-2mm}
\resizebox{\linewidth}{!}{
\renewcommand{\arraystretch}{0.85}
\begin{tabular}{lcccccccccccccc}
\toprule
\textbf{Sequence}
& \multicolumn{4}{c}{Morning}
& \multicolumn{4}{c}{Daytime}
& \multicolumn{4}{c}{Nighttime}
& \multicolumn{2}{c}{\textbf{Average}} \\
\cmidrule(lr){2-5}\cmidrule(lr){6-9}\cmidrule(lr){10-13}\cmidrule(l){14-15}
\textbf{ }
& \multicolumn{2}{c}{Campus} & \multicolumn{2}{c}{Residential}
& \multicolumn{2}{c}{Campus} & \multicolumn{2}{c}{Residential}
& \multicolumn{2}{c}{Campus} & \multicolumn{2}{c}{Residential}
& \multicolumn{2}{c}{ } \\
\cmidrule(lr){2-3}\cmidrule(lr){4-5}\cmidrule(lr){6-7}\cmidrule(lr){8-9}\cmidrule(lr){10-11}\cmidrule(l){12-13}
\textbf{Length}
& \multicolumn{2}{c}{7037m} & \multicolumn{2}{c}{3090m}
& \multicolumn{2}{c}{7071m} & \multicolumn{2}{c}{3151m}
& \multicolumn{2}{c}{6970m} & \multicolumn{2}{c}{3620m}
& \multicolumn{2}{c}{5157m} \\
\cmidrule(lr){2-3}\cmidrule(lr){4-5}\cmidrule(lr){6-7}\cmidrule(lr){8-9}\cmidrule(lr){10-11}\cmidrule(l){12-13}\cmidrule(l){14-15}
\textbf{Metric}
& $t_{\mathrm{apm}}$ & $CR$ & $t_{\mathrm{apm}}$ & $CR$
& $t_{\mathrm{apm}}$ & $CR$ & $t_{\mathrm{apm}}$ & $CR$
& $t_{\mathrm{apm}}$ & $CR$ & $t_{\mathrm{apm}}$ & $CR$
& $t_{\mathrm{apm}}$ & $CR$ \\
\midrule
\textbf{Without Loop-Closing} \\
\hspace{3mm} \monoicon~SVO 2.0 \cite{Forster17troSVO}
& --  & 0\% & -- & 0\%
& -- & 0\% & -- & 0\%
& -- & 0\% & -- & 0\%
& -- & 0\% \\
\hspace{3mm} \monoicon~DytanVO \cite{shen2023dytanvo}
& \textbf{.0266} & 100\% & .0439 & 100\%
& .0443 & 100\% & .0559 & 100\%
& .0569 & 100\% & .0538 & 100\%
& .0469 & 100\% \\
\hspace{3mm} \stereoicon~TartanVO \cite{tartanvo2020corl}
& .0379 & 100\% & .0492 & 100\%
& .0338 & 100\% & .0497 & 100\%
& .0421 & 100\% & .0397 & 100\%
& .0421 & 100\% \\
\rowcolor{ourgray}
\hspace{3mm} \stereoicon~\textbf{LST-SLAM (Ours)}
& .0329 & 100\% & \textbf{.0311} & 100\%
& \textbf{.0301} & 100\% & \textbf{.0399} & 100\%
& \textbf{.0297} & 100\% & \textbf{.0347} & 100\%
& \textbf{.0331} & 100\% \\
\midrule
\textbf{With Loop-Closing} \\
\hspace{3mm} \monoicon~MASt3R-SLAM \cite{murai2025mast3r}
& -- & 5.6\% & -- & 27.7\%
& -- & 8.9\% & -- & 12.2\%
& -- & 4.3\% & -- & 17.3\%
& -- & 12.7\% \\
\hspace{3mm} \stereoicon~ORB-SLAM3 \cite{campos2021orb}
& -- & 80.3\% & -- & 81.3\%
& -- & 84.1\% & .0136 & 100\%
& -- & 61.3\% & -- & 88.0\%
& -- & 72.4\% \\
\hspace{3mm} \stereoicon~OpenVSLAM \cite{sumikura2019openvslam}
& -- & 68.3\% & .0201 & 100\%
& -- & 87.1\% & .0179 & 100\%
& -- & 28.8\% & -- & 84.1\%
& -- & 77.9\% \\
\hspace{3mm} \stereoicon~OV$^{2}$SLAM \cite{ferrera2021ov}
& .0157 & 100\% & -- & 61.3\%
& .0227 & 100\% & .0189 & 100\%
& .0276 & 100\% & .0264 & 100\%
& .0227 & 93.6\% \\
\hspace{3mm} \stereoicon~AirSLAM \cite{xu2025airslam}
& .0243 & 100\% & .0627 & 100\%
& .0222 & 100\% & .0633 & 100\%
& .0222 & 100\% & .0583 & 100\%
& .0422 & 100\% \\
\hspace{3mm} \stereoicon~DROID-SLAM \cite{teed2021droid}
& .0132 & 100\% & .0531 & 100\%
& .0146 & 100\% & .0441 & 100\%
& .0154 & 100\% & .0438 & 100\%
& .0307 & 100\% \\
\hspace{3mm} \stereoicon~VINS-Fusion \cite{qin2019general}
& .0105 & 100\% & .0143 & 100\%
& .0166 & 100\% & .0167 & 100\%
& .0429 & 100\% & .0132 & 100\%
& .0203 & 100\% \\
\rowcolor{ourgray}
\hspace{3mm} \stereoicon~\textbf{LST-SLAM (Ours)}
& \textbf{.0067} & 100\% & \textbf{.0116} & 100\%
& \textbf{.0116} & 100\% & \textbf{.0068} & 100\%
& \textbf{.0131} & 100\% & \textbf{.0115} & 100\%
& \textbf{.0102} & 100\% \\
\bottomrule
\multicolumn{12}{l}{
\quad \monoicon~\hspace{.5mm} Monocular input.\quad
\stereoicon~\hspace{.5mm} Stereo input.} \\
\end{tabular}
}
\vspace{-15pt}
\label{tab:ate}
\end{table*}

\subsection{Thermal SLAM in Large-Scale Dynamic Environments}
\label{Method: Robustness and Accuracy Comparison of SLAM systems}

\noindent\textbf{Datasets \& Baseline}\; The MS$^{2}$ dataset provides synchronized stereo-thermal images from autonomous-driving scenes across multiple times of day, providing a kilometer-scale all-weather benchmark for stereo thermal SLAM \cite{shin2023deep}. We selected six large-scale sequences with loop closures, collected at different times and under different illumination conditions for the evaluation. Besides, we ran some of the most commonly used and well-known odometry and SLAM systems to compare the localization accuracy and robustness of our LST-SLAM with and without loop closure.

\begin{figure}[t]
  \centering
  \captionsetup{font=small}
  \includegraphics[width=\linewidth]{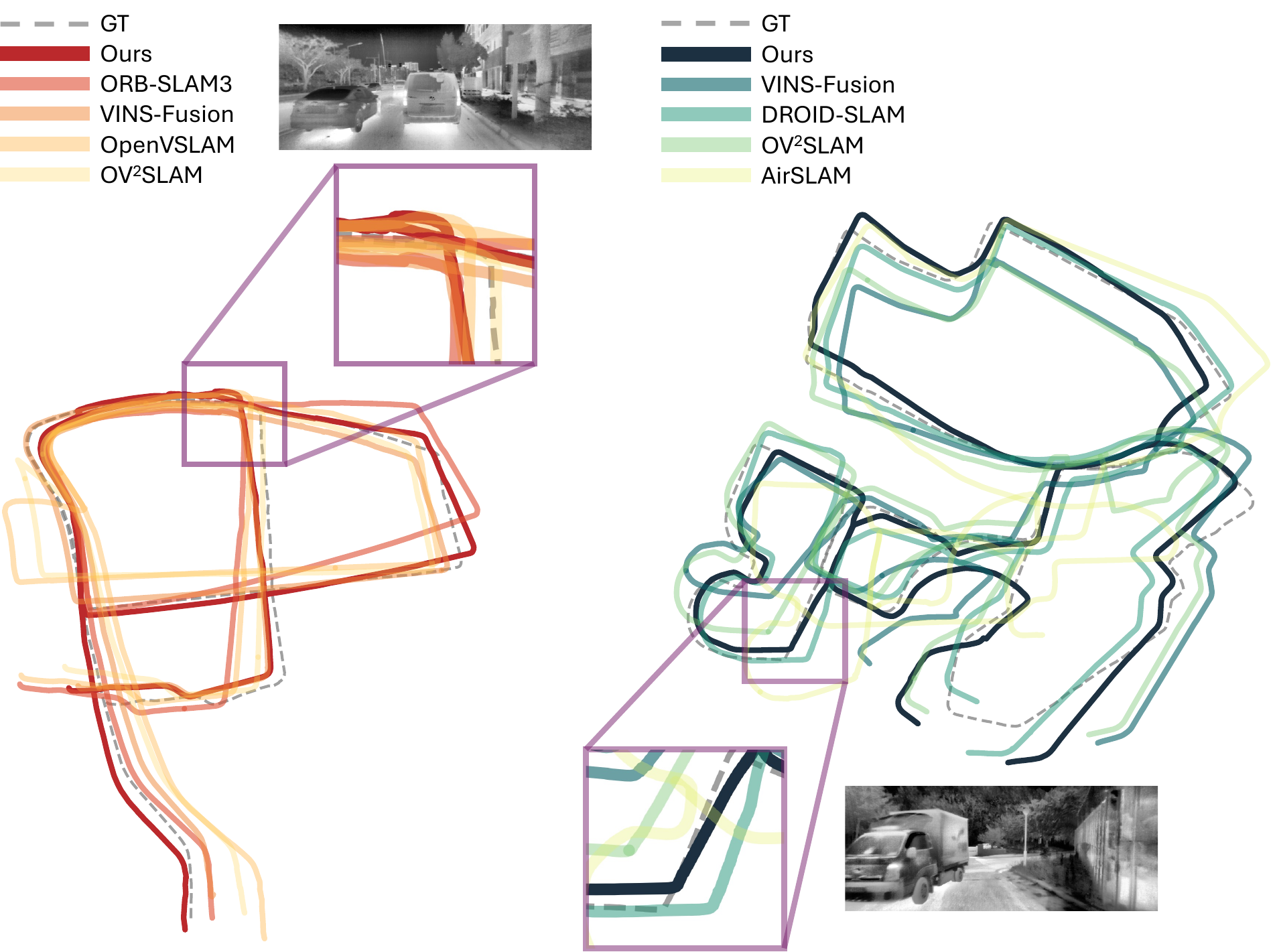}
  \caption{The qualitative results. \textbf{Left:} Daytime Residential. \textbf{Right:} Morning Campus. We visualize the trajectories of the top five most accurate methods and the ground truth trajectory for each scene.}
  \label{fig:traj}
  \vspace{-15pt}
\end{figure}

\noindent\textbf{Results \& Analysis}\; Since the systems incorporate a loop detection component, we adopt the absolute trajectory error (ATE) as the evaluation metric. The root mean square error (RMSE) is computed using \texttt{evo} \cite{grupp2017evo}. As this is the first work to evaluate kilometer-scale thermal SLAM datasets, we additionally introduce two metrics: $t_{\mathrm{apm}} = \mathrm{ATE}_{\mathrm{RMSE}}/L_{\mathrm{gt}}$, $CR = L_{\mathrm{align}}/L_{\mathrm{gt}}$, where $L_{\mathrm{gt}}$ is the total ground-truth trajectory length and $L_{\mathrm{align}}$ is the length of the scale-aligned estimated trajectory overlapping with the ground truth. For each method, we ran 10 iterations and selected the best trajectory for evaluation.

The quantified results are presented in \tref{tab:ate}. For $t_{\mathrm{apm}}$, we only calculate the average where all sequences are 100\% complete. For the comparison without loop detection, our method outperforms other VO methods. The SVO results show that traditional monocular odometry methods struggle to initialize in thermal environments. As daytime thermal images resemble grayscale images, our method achieved the highest accuracy in all sequences except for one captured in the morning. The average $t_{\mathrm{apm}}$ of LST-SLAM is 21.4\% lower than the second-best system, \ie TartanVO. For the comparison with loop detection, our method exhibits substantially higher robustness than other feature-based SLAM systems such as ORB-SLAM3, OpenVSLAM, and AirSLAM. In particular, compared to AirSLAM, which already demonstrates strong robustness, our approach achieves 75.8\% lower localization error. Moreover, relative to the deep learning-based DROID-SLAM and the optical flow-based VINS-Fusion, it achieves 66.8\% and 49.8\% lower error, respectively. Overall, LST-SLAM outperforms other representative systems under both loop-enabled and loop-disabled conditions. \fref{fig:traj} demonstrates that our method achieves superior pose estimation accuracy and generates smoother trajectories.

\subsection{Ablation Study}
\label{Method: Ablation Study on DST-SLAM}
We evaluated the contribution of each module using four variants:
\begin{itemize}
\item \textit{w/o STP:} replaces STP with SuperPoint for thermal feature extraction;
\item \textit{w/o Loop:} removes loop closure and global optimization;
\item \textit{w/o Seg:} excludes the dynamic object filtering module;
\item \textit{w/o DT:} replaces dual-level tracking with the motion model and BoW-guided projection from ORB-SLAM3.
\end{itemize}
%In \tref{tab:AblationStudy}, we evaluate the contribution of each of the four modules in our system. \textit{w/o STP} replaces the STP network with SuperPoint for feature extraction; \textit{w/o Loop} removes loop closure and global optimization; \textit{w/o Seg} excludes the dynamic object filtering module; and \textit{w/o DT} substitutes our dual-level tracking with the motion model and BoW-guided projection from ORB-SLAM3. The results for Campus and Residential are averaged over the three time periods.

%The experimental results in the table show that directly using SuperPoint substantially reduces robustness, making it infeasible to complete the accuracy evaluation. Compared with the \textit{w/o Loop}, \textit{w/o Seg}, and \textit{w/o DT} variants, our full DST-SLAM achieves average performance gains of 224.5\%, 46.1\%, and 33.3\%, respectively, clearly demonstrating the effectiveness of the proposed modules.

Results, averaged over three time periods for Campus and Residential sequences, are shown in \tref{tab:AblationStudy}. Directly substituting SuperPoint significantly degrades robustness, preventing the successful evaluation of accuracy. Compared with \textit{w/o Loop}, \textit{w/o Seg}, and \textit{w/o DT}, the full LST-SLAM achieves 69.2\%, 31.5\%, and 25.0\% lower localization error, respectively, demonstrating the effectiveness of each module.

\begin{table}[t]
\centering
\captionsetup{font=small}
\caption{Performance comparison of different ablation setups. We present the average results of six sequences in MS$^{2}$ datasets.}
\scriptsize
\setlength{\tabcolsep}{6pt}
\renewcommand{\arraystretch}{0.85}
\resizebox{\linewidth}{!}{
\begin{tabular}{lcccccc}
\toprule
\textbf{Method} 
& \multicolumn{2}{c}{Campus} 
& \multicolumn{2}{c}{Residential} 
& \multicolumn{2}{c}{Average} \\
\cmidrule(lr){2-3} \cmidrule(lr){4-5} \cmidrule(lr){6-7}
& $t_{\mathrm{apm}}$ & $CR$ & $t_{\mathrm{apm}}$ & $CR$ & $t_{\mathrm{apm}}$ & $CR$ \\
\midrule
w/o STP   & --     & 76.8\% & .0184 & 100\% & --     & 88.4\% \\
w/o Loop  & .0309 & 100\%   & .0352 & 100\% & .0331 & 100\% \\
w/o Seg   & .0140 & 100\%   & .0159 & 100\% & .0149 & 100\% \\
w/o DT    & .0127 & 100\%   & .0144 & 100\% & .0136 & 100\% \\
\midrule
\textbf{LST-SLAM} & \textbf{.0105} & \textbf{100\%} 
                  & \textbf{.0099} & \textbf{100\%} 
                  & \textbf{.0102} & \textbf{100\%} \\
\bottomrule
\end{tabular}}
\label{tab:AblationStudy}
\vspace{-15pt}
\end{table}

\section{Conclusion}

This work proposes a stereo thermal SLAM system designed for localization and mapping in dynamic, kilometer-scale thermal environments. It leverages self-supervised learning for thermal feature extraction and employs dual-level tracking for robust feature matching. Together with online iBoW loop detection and global optimization, our approach achieves superior localization accuracy and robustness compared to existing representative VO and SLAM systems. In future work, we aim to further improve the efficiency of LST-SLAM by developing a more lightweight thermal feature extractor. We also plan to explore deep learning-based feature matching strategies to enhance system robustness and to extend the framework with additional sensing modalities such as IMUs.


%%%%%%%%%%%%%%%%%%%%%%%%%%%%%%%%%%%%%%%%%%%%%%%%%%%%%%%%%%%%%%%%%%%%%%%%%%%%%%%%

%%%%%%%%%%%%%%%%%%%%%%%%%%%%%%%%%%%%%%%%%%%%%%%%%%%%%%%%%%%%%%%%%%%%%%%%%%%%%%%%

%%%%%%%%%%%%%%%%%%%%%%%%%%%%%%%%%%%%%%%%%%%%%%%%%%%%%%%%%%%%%%%%%%%%%%%%%%%%%%%%
\bibliographystyle{ieeetr} %ieeetr国际电气电子工程师协会期刊
\bibliography{main} % ref就是之前建立的ref.bib文件的前缀

\end{document}